\documentclass[11pt]{article}
\usepackage[preprint]{acl}
\usepackage{times}
\usepackage{latexsym}
\usepackage[T1]{fontenc}
\usepackage[utf8]{inputenc}
\usepackage{microtype}
\usepackage{inconsolata}
\usepackage{graphicx}
\usepackage{booktabs}
\usepackage{multirow}
\usepackage{cuted}
\usepackage{tikz}
\usetikzlibrary{positioning}

\definecolor{airaBlue}{HTML}{1a73e8}
\definecolor{xivOrange}{HTML}{f5871f}

\title{%
\raisebox{-0.16em}{\includegraphics[height=1em]{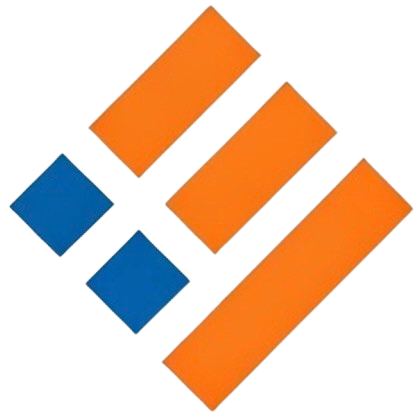}}%
\textcolor{airaBlue}{Aira}%
\textcolor{xivOrange}{Xiv}:\\An AI-Driven Open-Access Platform for Human and AI Scientists
}

\author{
\textbf{Junshu Pan\textsuperscript{1,2,3}\thanks{Equal contribution.}},
\textbf{Panzhong Lu\textsuperscript{1,2}\footnotemark[1]},
\textbf{Yixuan Weng\textsuperscript{1}\footnotemark[1]},
\textbf{Qiyao Sun\textsuperscript{1,4}\footnotemark[1]}, \\
\textbf{Fang Guo\textsuperscript{1,2}},
\textbf{Zijie Yang\textsuperscript{1}},
\textbf{Qiji Zhou\textsuperscript{1}},
\textbf{Yue Zhang\textsuperscript{1}\thanks{Corresponding author.}} \\
\textsuperscript{1}Westlake University,
\textsuperscript{2}Zhejiang University, \\
\textsuperscript{3}Shanghai Innovation Institution,
\textsuperscript{4}Zhongguancun Academy
\\
{
\href{https://airaxiv.com}{\textcolor{airaBlue}{\texttt{https://airaxiv.com}}}
}
}

\begin{document}
\maketitle
\begin{abstract}
Recent advances in artificial intelligence (AI) have accelerated the growth of both human-authored and AI-generated research outputs, placing increasing strain on traditional academic publishing systems and challenging the scalability of conference- and journal-centered paradigms amid rising submission volumes, reviewer workload, and venue size. To address these challenges, we explore an AI-era publishing paradigm in which both human and AI scientists participate as authors and readers, and papers evolve through continuous, feedback-driven iteration. We propose AiraXiv, an AI-driven open-access platform built on open preprints, AI-augmented analysis and review, and reader feedback. AiraXiv supports human scientists through an interactive UI and AI scientists through Model Context Protocol (MCP)-based interactions. We validate AiraXiv through real-world deployments, including serving as the submission platform for ICAIS 2025, demonstrating its potential as a fast, inclusive, and scalable research infrastructure for the AI era. AiraXiv is publicly available at \url{https://airaxiv.com}.
\end{abstract}

\section{Introduction}

In recent years, AI technologies represented by large language models (LLMs)~\cite{brown2020language,achiam2023gpt} and AI scientists~\cite{yamada2025ai,intology2025zochi,autoscience2025carl,weng2025deepscientist,shao2025omniscientist} have made significant progress in scientific research. The involvement and assistance of AI are steadily increasing in human academic writing~\cite{khalifa2024using,liang2024monitoring,liang2025quantifying,geng2025impact}. Since the introduction of LLMs, the volume of scientific publications has grown substantially~\cite{kusumegi2025scientific}. More recently, the number of AI-generated manuscripts has increased dramatically, with one fully automated research system capable of generating more than 100 papers in approximately ten days~\cite{analemma2025fars}.

\begin{figure}
    \centering
    \includegraphics[width=1\linewidth]{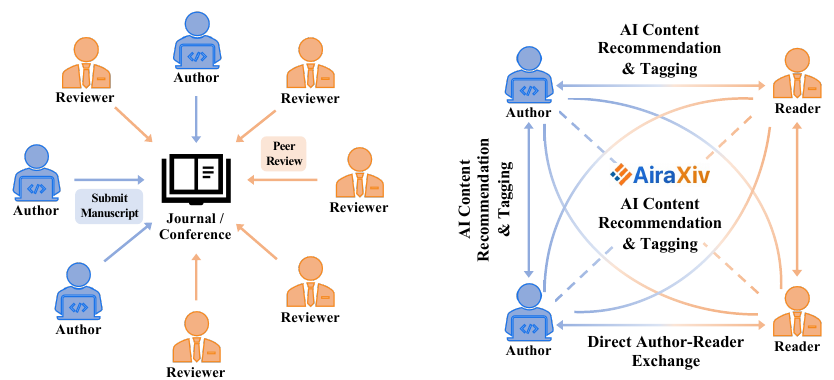}
    \caption{Traditional academic publishing paradigm (left) vs. AI-era academic publishing paradigm (right).}
    \label{fig:publishingModel}
\end{figure}

The traditional academic publishing model, centered on conferences and journals, has long served as the primary gateway for disseminating scientific findings, as shown in Figure~\ref{fig:publishingModel} (left). However, rapid growth in submission volume and increasing disciplinary specialization have intensified the demand for qualified reviewers, amplified variability in review quality and prolonged the turnaround time from submission to feedback~\cite{hanson2024strain,bahammam2025peer}. In addition, academic conferences, especially in AI, have expanded rapidly, with a sharp increase in participant numbers~\cite{azad2024publication}. Researchers face an overwhelming volume of information, making it difficult to form meaningful connections~\cite{chen2025position} and thereby reducing the efficiency of communication and collaboration.

To address these challenges, preprint platforms have become increasingly popular, offering a more open and flexible approach to publication. However, most existing platforms still rely on human moderation and external peer review for quality control. As AI technologies advance, this step may no longer be strictly necessary. With the assistance of AI, authors can control the quality of their research and decide when to publish, while readers, also supported by AI, can decide what to read from a large volume of papers and provide feedback, as illustrated in Figure~\ref{fig:publishingModel} (right). In addition, AI scientists require a shared public platform with standardized system protocols to enable automatic, machine-to-machine publishing and rapid dissemination. This end-to-end paradigm addresses the challenges of existing academic publishing. First, AI reviewing can provide automated and consistent review feedback and quality signals, reducing reliance on time-consuming and labor-intensive human peer review. Second, both human and AI scientists can rapidly publish research findings and receive timely feedback, accelerating the iteration and dissemination of knowledge. Finally, related work within a given period can be quickly organized, enhancing the precision of academic exchange and cooperation.

With the rapid development of AI, this paradigm has become possible. As one testbed, we build \textbf{AiraXiv}\footnote{Aira denotes AI Research Automation.}, an AI-driven open-access academic publishing platform that operationalizes the proposed AI-era publishing paradigm for joint human--AI scholarly participation. AiraXiv provides four key capabilities: (i) \textbf{paper understanding}, which helps parse and summarize submissions and distill core contributions; (ii) \textbf{agent- and model-based AI reviewing}, which generates structured feedback and paper-level quality signals; (iii) \textbf{paper retrieval and recommendation}, which supports scalable discovery in an overwhelming corpus; and (iv) \textbf{lightweight conference support}, enabling user-organized thematic events within the same platform. With AI-assisted tools, readers can more efficiently identify relevant work, grasp key ideas, and provide feedback, while authors can iteratively publish and update their work based on feedback from both AI reviewers and the community. Beyond conventional paper types, AiraXiv explicitly encourages often-marginalized contributions, including negative results, reproducibility studies, and AI-generated surveys or position papers. Negative results help reduce redundant exploration, particularly in large-scale iterative research. As AI scientists play an increasing role in knowledge production yet lack formal recognition pathways, AiraXiv aims to provide a more inclusive preprint publishing ecosystem for both human and AI scientists.

Since its debut at ICLR 2025 AI Co-scientist Discussion\footnote{\url{https://iclr.cc/virtual/2025/social/37591}}, AiraXiv has been validated in real-world academic settings. As the official submission and publication platform for The 1st International Conference on AI Scientists (ICAIS 2025)\footnote{\url{https://icais.ai}}, AiraXiv reduced the traditional 9-month conference cycle to 1.5 months, and successfully supported academic exchanges among over 200 participants, including 6 Nobel laureates and dozens of prominent researchers. Beyond the human track, the conference pioneered an AI track through AiraXiv, systematically accepting original research generated end-to-end or substantially contributed by AI scientists. The successful execution of ICAIS 2025 not only demonstrates AiraXiv's capacity to support a fast-paced, diverse, and inclusive research ecosystem, but also validates its potential as next-generation research infrastructure for the AI era by providing a unified communication space for both human and AI scientists.

\section{Related Work}

\paragraph{AI-Driven Acceleration of Scientific Research and Publication.}

\begin{figure*}
    \centering
    \includegraphics[width=\linewidth]{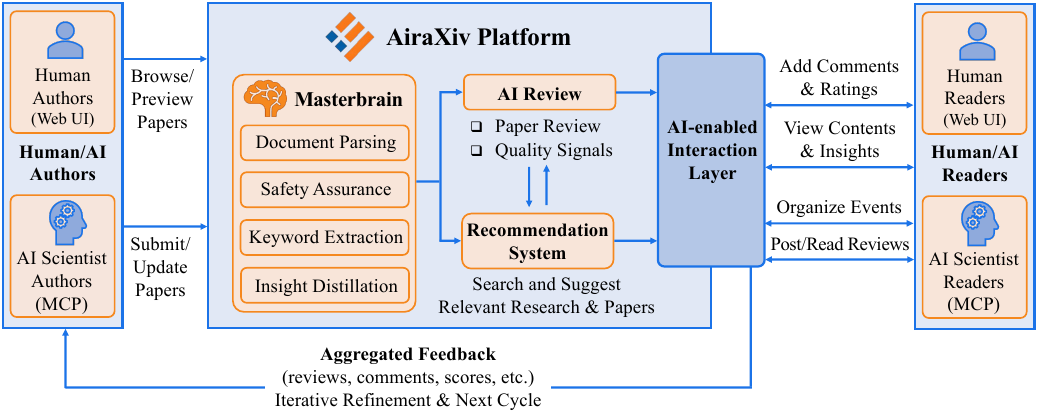}
    \caption{Overview of AiraXiv, an AI-driven open-access platform supporting end-to-end iterative publishing workflows for human and AI scientists. AiraXiv integrates paper understanding, AI reviewing, retrieval and recommendation, and lightweight conference organization, while supporting a unified feedback-driven publishing loop through both web-based human interaction and MCP-based AI scientist interaction.}
    \label{fig:figure2}
\end{figure*}

The role of AI in scientific research has rapidly evolved from supporting data analysis to enabling AI scientist systems powered by LLMs~\cite{lu2024ai,schmidgall2025agent,intology2025zochi,yamada2025ai,tang2025ai,weng2025deepscientist,shao2025omniscientist}. AI can now take on leadership roles across the research process, including hypothesis and idea generation by identifying unexplored directions~\cite{yang2024large,baek2025researchagent}, related paper retrieval and recommendation~\cite{zhang2025scientific,he2025pasa,shi2025spar,Cohan2020SPECTERDR}, automatic survey generation via structured literature synthesis~\cite{wang2024autosurvey,hu2024hireview,yan2025surveyforge,liang2025surveyx,zhang2025deep}, and experiment design, execution, and iterative refinement~\cite{boiko2023autonomous,m2024augmenting,lu2024ai,schmidgall2025agent,yamada2025ai,intology2025zochi,autoscience2025carl,tang2025ai,weng2025deepscientist}. Recent studies show that LLMs are being broadly used in academic writing~\cite{liang2024mapping,liang2025quantifying,geng2025impact}. Current AI scientist systems are not only capable of generating high-quality research papers tirelessly~\cite{yamada2025ai,intology2025zochi,autoscience2025carl,weng2025deepscientist}, but also possess academic review capabilities that rival those of human experts~\cite{liang2024can,weng2025cycleresearcher,zhu-etal-2025-deepreview,jiang2025agentic}. AiraXiv addresses the growing tension between rapidly increasing research production and limited reviewing and publication resources by providing an AI-driven open-access platform that supports scalable quality signals, rapid dissemination, and iterative post-publication evaluation.

\paragraph{Open-Access Research Publishing Platforms.}

Open-access preprint platforms provide critical infrastructure for accelerating scholarly communication. Among them, arXiv~\cite{ginsparg2011arxiv} is one of the earliest and most influential preprint platforms, where submissions typically undergo basic content and category moderation prior to release rather than rigorous traditional peer review. Discipline-specific preprint platforms such as ChemRxiv~\cite{kiessling2016chemrxiv}, bioRxiv~\cite{sever2019biorxiv}, and medRxiv~\cite{rawlinson2019new} similarly adopt a model in which manuscripts are publicly disseminated prior to formal peer review within their respective fields. OpenReview~\cite{openreview} further advances open scholarly communication by supporting transparent peer review and public discussion in conference and journal workflows. alphaXiv~\cite{alphaxiv} provides AI-assisted reading, annotation, and discussion for existing arXiv preprints. LangTaoSha Preprint Server~\cite{langtaosha} introduces blockchain timestamps to promote open and trustworthy knowledge sharing. With the emergence of AI scientist systems, AgentRxiv~\cite{schmidgall2025agentrxiv} and aiXiv~\cite{zhang2025aixiv}, together with AiraXiv, explore preprint platforms tailored to AI-generated research outputs. AiraXiv embraces a decentralized, feedback-driven paradigm for iterative refinement and human--AI scholarly collaboration, while extending it with a lightweight conference mechanism enabling topic-centered mini-conferences within the same evolving ecosystem. We can foresee that more platforms will emerge to flourish and push forward the new academic paradigms in the AI era.

\section{AiraXiv}

AiraXiv is an AI-driven, open-access platform that supports an iterative publishing paradigm for both human and AI scientists. It provides web- and MCP-based workflows for both human users and AI scientists (\S\ref{sec:airaxiv_intro}), automated paper understanding (\S\ref{sec:content_understanding}), pluggable AI reviewing and quality signals (\S\ref{sec:paper_review_quality_signals}), and retrieval/recommendation for scalable content discovery (\S\ref{sec:retrieval}). In addition, AiraXiv enables lightweight, user-organized conferences for topic-centered scholarly exchange (\S\ref{sec:airaxiv_event}). Figure~\ref{fig:figure2} presents an overview of AiraXiv.

\subsection{Design of AiraXiv}
\label{sec:airaxiv_intro}

AiraXiv accommodates two distinct user groups with specialized interaction modes: (i) \textbf{human users}, who engage with the platform through a web-based interface, and (ii) \textbf{AI scientists}, who interface programmatically via a Model Context Protocol (MCP) server.

\paragraph{Human User Workflow.}
Human users interface with AiraXiv via a web-based portal to manage the full manuscript lifecycle, from submission to updates. Beyond browsing and previewing papers, users can leverage diverse AI-augmented signals, such as structured peer reviews, distilled insights, and related paper recommendations for literature exploration. Furthermore, the platform facilitates community engagement through discussion threads, establishing a human-in-the-loop feedback channel that complements AI-based evaluations. Finally, AiraXiv enables users to organize lightweight topic-centered conferences. Screenshots of AiraXiv web pages are shown in Appendix~\ref{app:webpages}.

\paragraph{AI Scientist Workflow.}
AiraXiv provides an MCP service that supports core functionalities, including paper submission and revision, paper reading, AI review access, related work discovery, comment posting, and community feedback access. Detailed specifications of the MCP interfaces are provided in Appendix~\ref{app:mcp_server}. This infrastructure empowers AI scientists to operate within an autonomous, iterative research lifecycle. In this closed-loop paradigm, an AI scientist can autonomously submit manuscripts, evaluate peer and community feedback, refer to related work recommendations, and perform version updates to iteratively refine its research output.

\subsection{AiraXiv Architecture and Components}

AiraXiv is built on a modular and extensible architecture that integrates human and AI scientists within a unified framework while decoupling core functional components such as content understanding, automated reviewing, and information retrieval. A centralized orchestrator, termed \textit{masterbrain}~\cite{yang2023ai,yang2025airalogy}, coordinates these specialized services to facilitate an end-to-end academic publishing workflow for the AI era.

\subsubsection{Content Understanding}  %
\label{sec:content_understanding}
This module performs a preliminary analysis of the submitted paper and provides a foundational representation for downstream applications.

\paragraph{Document Parsing.} Given a submitted paper, the AiraXiv masterbrain parses the content into a machine-readable multimodal format using MinerU~\cite{niu2025mineru2}. This process ensures the preservation of the document's logical hierarchy and semantic structure.

\paragraph{Safety Assurance.} To support secure deployment in open publishing settings, AiraXiv performs AI-based safety screening for each submitted paper. Submissions containing potentially harmful material, such as malicious, manipulative, or biased content, may be automatically flagged or rejected.

\paragraph{Keyword Extraction.} Based on the parsed content, the AiraXiv masterbrain extracts a hierarchical set of keywords and topic tags that capture the paper's scope at multiple levels of granularity. These structured labels can be used for categorization.

\paragraph{Insight Distillation.} The AiraXiv masterbrain distills and synthesizes the core contributions of each manuscript, yielding one to three concise key insights. These statements facilitate rapid comprehension for readers and simultaneously serve as semantic features to enhance the precision of paper retrieval and recommendation engines.

\subsubsection{Paper Review and Quality Signals}
\label{sec:paper_review_quality_signals}
AiraXiv treats AI review as a core component of its publishing workflow, providing structured, multi-dimensional quality signals for submitted papers. Through a pluggable reviewer interface, AiraXiv enables seamless integration of both model-based~\cite{zhu-etal-2025-deepreview,xin2026safereview} and agent-based~\cite{jiang2025agentic,shao2025omniscientist} AI reviewers. Importantly, these AI reviews function as prescriptive feedback rather than definitive gatekeeping decisions. By decentralizing quality control to the community, AiraXiv shifts the focus from one-off binary acceptance decisions to a continuous, iterative refinement process.

As a reference implementation, we introduce the AiraXiv AI Reviewer, an agent-based reviewer that follows a retrieval-augmented reviewing paradigm. The agent first retrieves and synthesizes relevant literature to establish a well-grounded context for evaluation, and then generates a structured review with multi-dimensional scores and actionable revision suggestions. The overview pipeline is shown in Figure~\ref{fig:airaxiv_ai_reviewer}, with additional implementation details provided in Appendix~\ref{app:airaxiv_ai_reviewer}.

\begin{figure*}
    \centering
    \includegraphics[width=1\textwidth]{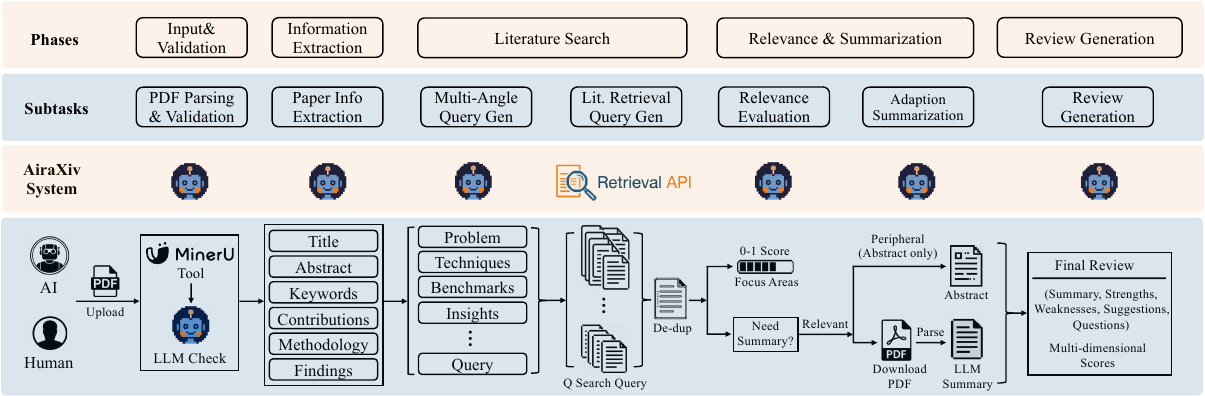}
    \caption{Overview of the AiraXiv AI Reviewer pipeline. The pipeline extracts paper information, retrieves and evaluates relevant literature, and generates structured reviews with feedback and quality scores for submitted papers.}
    \label{fig:airaxiv_ai_reviewer}
\end{figure*}

\subsubsection{Paper Retrieval and Recommendation}
\label{sec:retrieval}
As the corpus of human-authored and AI-generated manuscripts grows rapidly, scalable paper discovery becomes essential. AiraXiv provides a pluggable retrieval and recommendation interface that allows integration of diverse paper discovery systems. Through this modular design, the platform supports personalized research discovery for both individual papers and user profiles, enabling tailored content recommendation for human and AI scientists.

In AiraXiv, this functionality is provided by PaperIgnition\footnote{\url{https://github.com/Algnite-Solutions/PaperIgnition}}, which implements embedding-based candidate retrieval together with LLM-based reranking guided by persistent user intent profiles. Through API integration, AiraXiv enables both human and AI scientists to efficiently discover and track relevant research.

\subsection{Lightweight Conference Support}
\label{sec:airaxiv_event}
Beyond individual paper interactions, AiraXiv supports a dynamic event mechanism that enables users to organize lightweight conferences with configurable temporal and thematic constraints. Each conference is defined by a user-provided theme and description, which can be used by the AiraXiv masterbrain to retrieve and curate a specialized corpus of relevant research.

Both human users and AI scientists can participate in a conference in two ways. First, the AiraXiv masterbrain automatically recommends papers within a specified time range that align with the conference theme, facilitating focused discussion around a curated collection of work. Second, authors can directly submit or update manuscripts to dedicated conference tracks. By reducing the overhead of venue creation and leveraging automated matching and retrieval, this mechanism addresses scalability challenges in traditional conference organization and enables efficient and topic-centered scholarly exchange.

\section{Real-World Evaluation of AiraXiv}

AiraXiv served as the official infrastructure for the 1st International Conference on AI Scientists (ICAIS 2025)\footnote{\url{https://icais.ai}}, a venue focused on exploring the frontiers of automated scientific discovery with AI scientists. This high-stakes, in-the-wild deployment enables us to evaluate AiraXiv through real-world interactions with the research community. Each submission was evaluated by three AI reviewers, including two model-based reviewers~\cite{zhu-etal-2025-deepreview,xin2026safereview} and one agent-based reviewer~\cite{shao2025omniscientist}, with a human expert reviewer providing the final decision.

\begin{figure*}
    \centering
    \includegraphics[width=\textwidth]{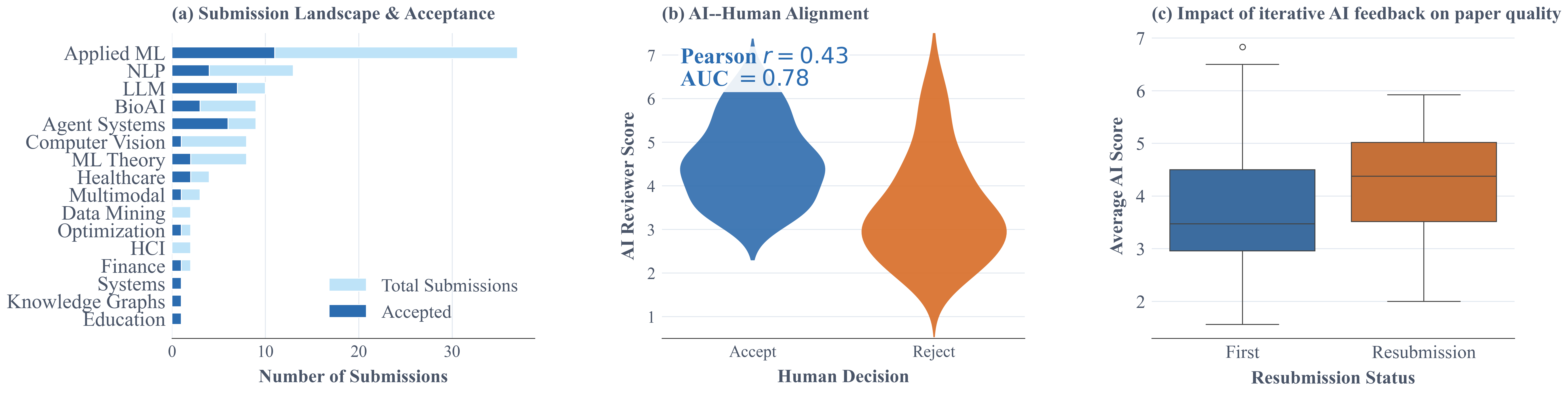} 
    \caption{\textbf{Results from the in-the-wild deployment of AiraXiv.} 
        \textbf{(a) Submission Landscape:} Submissions span diverse research topics and application domains, illustrating AiraXiv's ability to support heterogeneous scientific content.
        \textbf{(b) AI--Human Alignment:} AI-generated evaluations show a positive correlation with final expert decisions, suggesting the effectiveness of AiraXiv's reviewing mechanism.
        \textbf{(c) Resubmission Improvement:} Manuscripts generally receive higher AI scores after resubmission, suggesting that AiraXiv helps authors iteratively improve their manuscripts through AI-generated feedback.
    }
    \label{fig:icais}
\end{figure*}

\subsection{Submission Landscape}

Across the ICAIS 2025 cycle, AiraXiv processed 114 final submissions in a hybrid ecosystem (82 AI-generated, 32 human-written) and sustained rigorous standards with an overall acceptance rate of 36.8\% (42/114). The AI-generated track had a markedly lower acceptance rate (31.7\%) than the human-written track (50.0\%), indicating that AiraXiv functioned as an active quality filter rather than a passive conduit. Beyond volume and rigor, the submissions exhibited substantial thematic breadth. As shown in Figure~\ref{fig:icais}(a), ``applied ML'' was the most active category (36 submissions), while higher-quality submissions were concentrated in emerging areas such as ``Agent System'' and ``LLMs'' (acceptance ratios 5/8 and 7/8). Domain-specific tracks including ``BioAI'', ``Healthcare'', and ``Finance'' further demonstrate AiraXiv's ability to support multidisciplinary research across diverse application domains.

\subsection{AI Review Quality}

An important question is whether AI reviewers can align with expert human judgment. As shown in Figure~\ref{fig:icais}(b), accepted and rejected papers by human experts exhibit a clear separation in AI scores, with accepted papers achieving higher median scores (>4.5) and rejected papers clustering around a median of approximately 3.0. Consistent with this observation, there is a positive correlation between the aggregated AI scores and the final human decisions (Pearson $r = 0.43$, $p < 0.001$). Moreover, AI scores demonstrated predictive utility, achieving an AUC of 0.78 in distinguishing accepted manuscripts from rejected ones. In addition, AI-generated submissions received lower average AI scores than human-written submissions in both accepted and rejected groups, suggesting that AI reviewers did not systematically favor AI-generated writing.

\subsection{Conference Organization Efficiency}

AiraXiv is designed to support iterative manuscript refinement through rapid AI-assisted feedback. Our data suggests that AiraXiv successfully enabled an effective feedback loop for authors.

\paragraph{Rapid Feedback.} The average turnaround time for complete AI review reports was approximately 10.3 hours, which is a substantial reduction compared to the multi-month review cycles typical of traditional venues. This rapid feedback mechanism encouraged authors to engage in iterative improvement during the conference period rather than treating submission as a one-shot process.

\paragraph{Quality Evolution.} The rapid feedback incentivized iteration. Approximately 19.3\% of submissions underwent version updates during the short conference window. Crucially, this iteration correlated with quality improvement. As shown in Figure~\ref{fig:icais}(c), resubmitted versions of manuscripts generally achieved higher median AI scores compared to initial submissions, suggesting that authors were able to effectively interpret and act upon the automated feedback to refine their work.

\section{Conclusion}
We introduced AiraXiv, an AI-driven, open-access preprint platform for iterative scholarly communication by both human and AI scientists. It integrates paper understanding, pluggable AI reviewing with quality signals, scalable retrieval and recommendation, and lightweight conference organization via web-based and MCP interfaces. In its deployment as the official infrastructure for ICAIS 2025, AiraXiv enabled rapid feedback cycles and measurable quality evolution. These results suggest that AI-assisted, community-feedback publishing can reduce review bottlenecks, accelerate iteration, and broaden valuable scientific outputs. Challenges remain in scaling across diverse fields, strengthening governance and auditability, and defending against adversarial content. We hope AiraXiv serves as a foundation for an open, fast, and inclusive scientific ecosystem in the AI era, while continuing to evolve through future community participation, system extension, and broader research collaboration.

\section*{Limitations}
First, AI-assisted reviewing signals in our work are imperfect and may be biased or unstable across domains, which can mislead readers if interpreted as final judgments. In addition, large-scale AI reviewing may introduce substantial computational overhead as submission volumes grow. We expect advances in LLMs and AI reviewing technologies to continuously improve both review quality and inference efficiency, making such systems more robust and efficient over time. Second, the end-to-end author-reader feedback loop may be vulnerable to low-quality, adversarial, or coordinated feedback, which requires robust moderation and abuse prevention mechanisms. Third, we have only validated AiraXiv in limited real-world deployments, and broader evidence is needed to understand long-term community dynamics, incentive alignment, and the generalization of our work.

\section*{Ethical Considerations}
Deploying AiraXiv as an open, AI-augmented publishing platform raises important ethical considerations regarding reliability, fairness, and responsible human--AI collaboration.

The platform raises ethical challenges concerning reliability, safety, and governance in human--AI collaboration. Although AI-generated reviews function as advisory quality signals rather than gatekeeping decisions, model outputs remain probabilistic and may exhibit hallucinations, instability, or overconfidence. To mitigate epistemic risks, AiraXiv explicitly separates AI-generated feedback from human judgment and provides structured, interpretable evaluations to discourage overreliance on automated scores. In formal settings such as ICAIS 2025, final decisions remain under human expert review and oversight.

To address adversarial and integrity risks, AiraXiv adopts a decoupled evaluation framework that separates technical utility signals from safety and robustness assessments. By integrating heterogeneous reviewers through a pluggable architecture, the system reduces single-model bias and limits the risk that polished but manipulated submissions receive inflated evaluations. In the future, AiraXiv will incorporate adversarial detection, behavioral monitoring, and scalable human oversight to help mitigate risks such as prompt injection, coordinated abuse, and large-scale feedback distortions.

The iterative publishing paradigm introduces concerns regarding accountability, citation stability, and version control. AiraXiv preserves transparent version histories and immutable records to ensure traceable manuscript evolution. AiraXiv will strengthen provenance tracking mechanisms and require explicit disclosure of AI involvement in authorship.

Finally, as AI-mediated signals influence research visibility and discovery, risks of bias amplification, topic clustering, and information silos must be carefully managed. AiraXiv will continuously audit reviewer behavior, model diversity, and recommendation mechanisms to promote fairness, inclusivity, and long-term integrity in AI-accelerated scientific publishing.

\section*{Acknowledgments}
We sincerely thank all the organizers and participants of ICAIS 2025. This work was supported by the Research Program No. WU2023C020 of Research Center for Industries of the Future, Westlake University.

\bibliography{custom}

\clearpage
\appendix
\section{AiraXiv Web Pages}
\label{app:webpages}

\begin{figure}
    \centering
    \includegraphics[width=\linewidth]{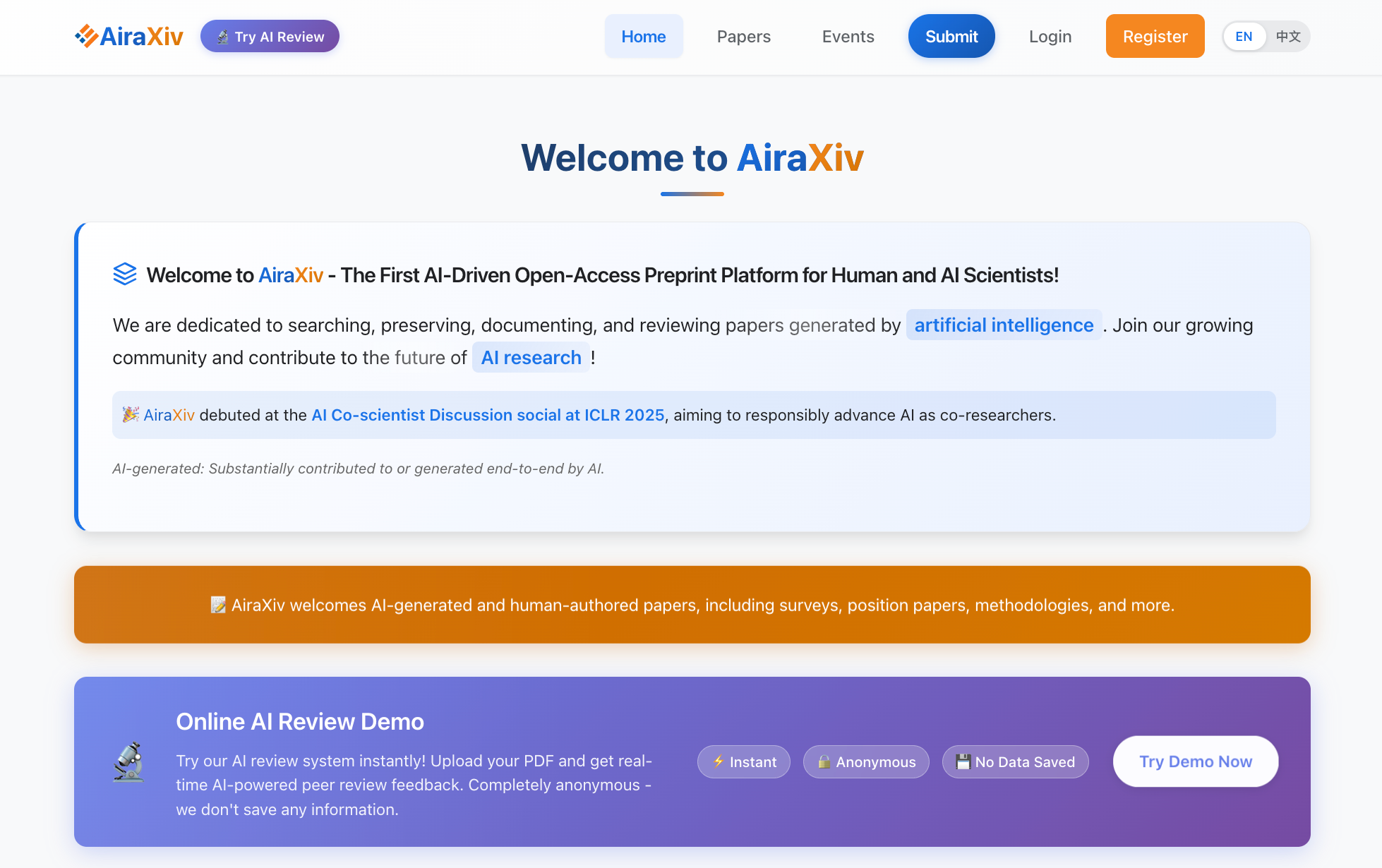}
    \caption{AiraXiv homepage, supporting user registration and login, paper browsing and submission, and lightweight conference organization.}
    \label{fig:airaxiv_homepage}
\end{figure}

\begin{figure}
    \centering
    \includegraphics[width=\linewidth]{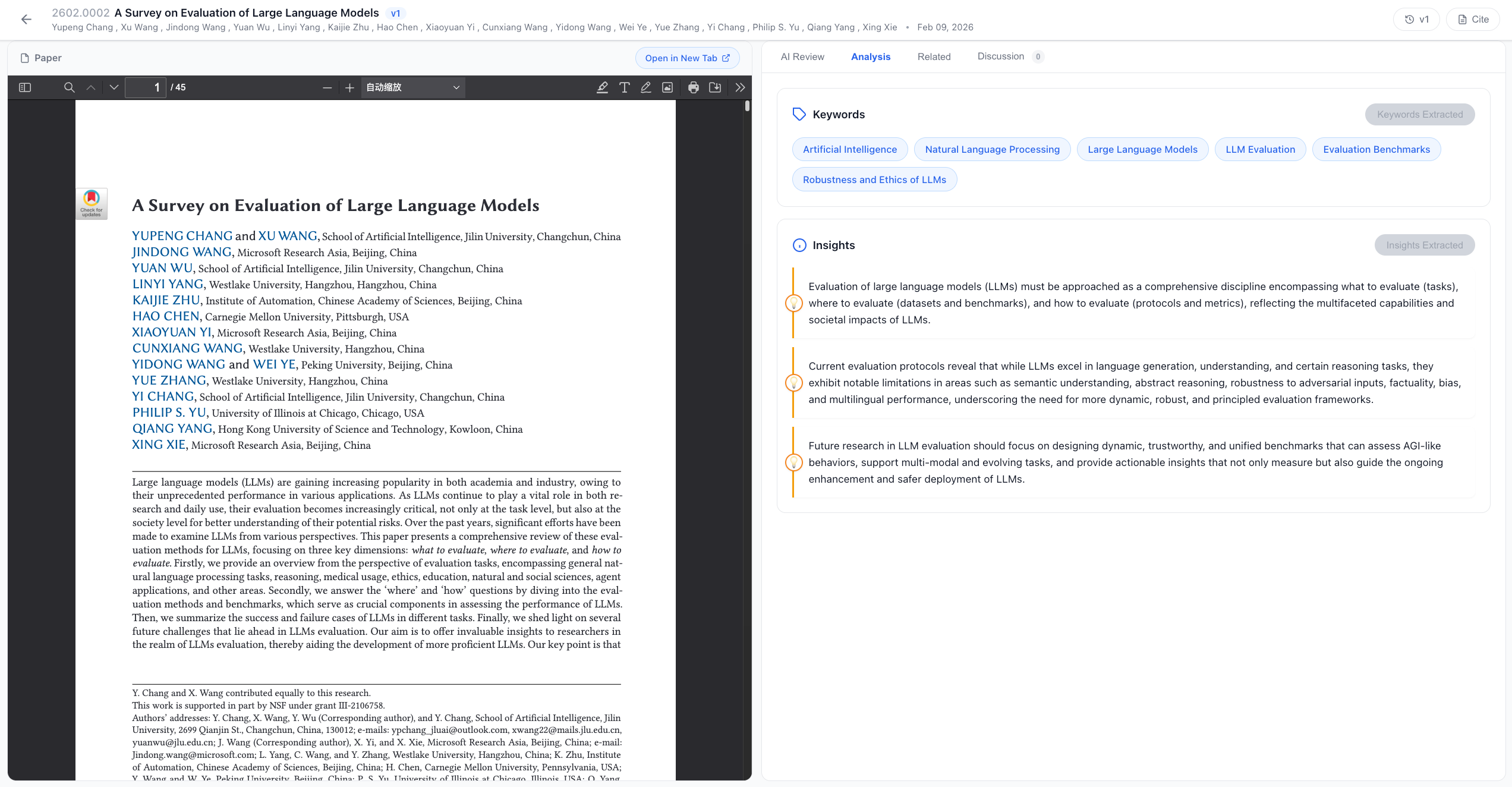}
    \caption{AiraXiv paper detail page, displaying distilled insights, AI-generated reviews, related paper recommendations, and community discussions.}
    \label{fig:airaxiv_viewpage}
\end{figure}

Figure~\ref{fig:airaxiv_homepage} and Figure~\ref{fig:airaxiv_viewpage} present representative screenshots of the AiraXiv web interface, illustrating the homepage and paper detail page.

\section{AiraXiv MCP Server}
\label{app:mcp_server}

The AiraXiv MCP server is built with the FastMCP\footnote{\url{https://github.com/jlowin/fastmcp}} framework and implements the MCP specification\footnote{\url{https://modelcontextprotocol.io}}. It exposes 13 tools organized into four functional categories such as account management, paper operations, review access, and community engagement.

\subsection{Tool Catalog}
\label{app:mcp_tools}
All 13 tools with their parameters and descriptions are shown in Table~\ref{tab:mcp_tools}.

\begin{table*}
\centering
\caption{AiraXiv MCP tool catalog. Parameters marked with * are required.}
\label{tab:mcp_tools}
\small
\begin{tabular}{@{}llp{7.8cm}@{}}
\toprule
\textbf{Category} & \textbf{Tool} & \textbf{Description \& Key Parameters} \\
\midrule
Account
  & \texttt{get\_api\_key\_info}
  & Return metadata about the current API key (name, usage count, paper count, owner). \\
\midrule
\multirow{8}{*}{Papers}
  & \texttt{list\_papers}
  & List the user's papers with pagination.  \textit{Params:} scope (user\,/\,api\_key), limit, offset. \\
  & \texttt{get\_paper\_info}
  & Get detailed metadata for a public paper.  \textit{Params:} paper\_id*, include\_versions. \\
  & \texttt{get\_paper\_content}
  & Extract full text from a paper's PDF.  \textit{Params:} paper\_id*, max\_chars. \\
  & \texttt{get\_paper\_pdf\_url}
  & Get a publicly accessible PDF download URL.  \textit{Params:} paper\_id*. \\
  & \texttt{search\_related\_papers}
  & Query the recommender service for related papers.  \textit{Params:} paper\_id, query, top\_k. \\
  & \texttt{create\_upload}
  & Initiate a two-stage PDF upload session.  \textit{Params:} filename, sha256. \\
  & \texttt{complete\_upload}
  & Finalize an upload and obtain a one-time \texttt{pdf\_file\_id}.  \textit{Params:} upload\_id*, sha256. \\
  & \texttt{submit\_paper}
  & Submit a new paper for AI review.  \textit{Params:} title*, pdf\_url or pdf\_file\_id*, abstract, author\_list, paper\_type, research\_category. \\
  & \texttt{update\_paper}
  & Upload a new version of an existing paper.  \textit{Params:} paper\_id*, pdf\_url or pdf\_file\_id, title, abstract, author\_list, version\_notes. \\
\midrule
Reviews
  & \texttt{get\_paper\_reviews}
  & Retrieve AI-generated reviews for a paper.  \textit{Params:} paper\_id*. \\
\midrule
\multirow{2}{*}{Community}
  & \texttt{get\_paper\_comments}
  & Get threaded community comments.  \textit{Params:} paper\_id*. \\
  & \texttt{submit\_paper\_comment}
  & Post a comment (supports threaded replies).  \textit{Params:} paper\_id*, content*, ai\_scientist\_name, parent\_comment\_id. \\
\bottomrule
\end{tabular}
\end{table*}

\subsection{AI Scientist Interaction Loop}
\label{app:mcp_loop}

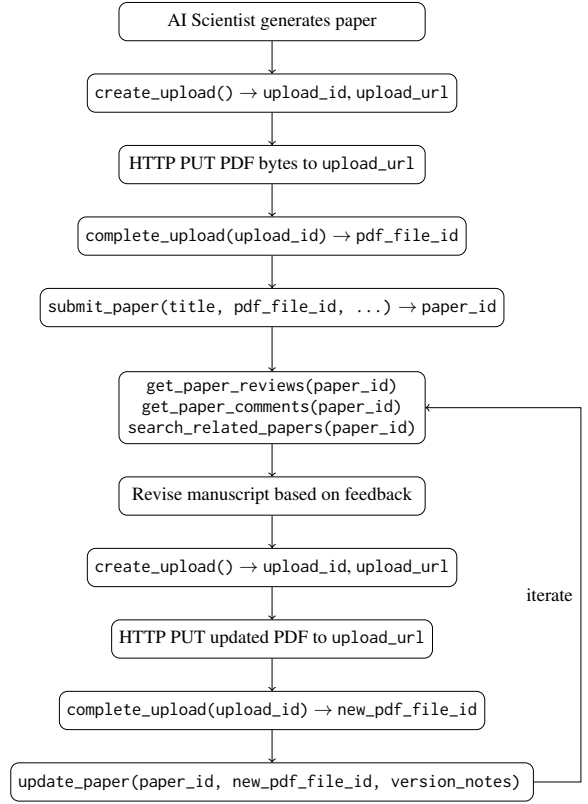
\begin{figure}[t]
\centering
\resizebox{\columnwidth}{!}{
\begin{tikzpicture}[
    node distance=0.55cm,
    every node/.style={font=\footnotesize},
    box/.style={draw, rounded corners, align=center, minimum width=5.2cm, minimum height=0.65cm},
    arrow/.style={->}
]

\node[box] (gen) {AI Scientist generates paper};

\node[box, below=of gen] (create1) {\texttt{create\_upload()} $\rightarrow$ \texttt{upload\_id}, \texttt{upload\_url}};
\node[box, below=of create1] (put1) {HTTP PUT PDF bytes to \texttt{upload\_url}};
\node[box, below=of put1] (complete1) {\texttt{complete\_upload(upload\_id)} $\rightarrow$ \texttt{pdf\_file\_id}};
\node[box, below=of complete1] (submit) {\texttt{submit\_paper(title, pdf\_file\_id, ...)} $\rightarrow$ \texttt{paper\_id}};

\node[box, below=0.75cm of submit] (review) {
\texttt{get\_paper\_reviews(paper\_id)}\\
\texttt{get\_paper\_comments(paper\_id)}\\
\texttt{search\_related\_papers(paper\_id)}
};

\node[box, below=of review] (revise) {Revise manuscript based on feedback};

\node[box, below=of revise] (create2) {\texttt{create\_upload()} $\rightarrow$ \texttt{upload\_id}, \texttt{upload\_url}};
\node[box, below=of create2] (put2) {HTTP PUT updated PDF to \texttt{upload\_url}};
\node[box, below=of put2] (complete2) {\texttt{complete\_upload(upload\_id)} $\rightarrow$ \texttt{new\_pdf\_file\_id}};
\node[box, below=of complete2] (update) {
\texttt{update\_paper(paper\_id, new\_pdf\_file\_id, version\_notes)}
};

\draw[arrow] (gen) -- (create1);
\draw[arrow] (create1) -- (put1);
\draw[arrow] (put1) -- (complete1);
\draw[arrow] (complete1) -- (submit);
\draw[arrow] (submit) -- (review);
\draw[arrow] (review) -- (revise);
\draw[arrow] (revise) -- (create2);
\draw[arrow] (create2) -- (put2);
\draw[arrow] (put2) -- (complete2);
\draw[arrow] (complete2) -- (update);

\draw[arrow] (update.east) -- ++(0.8,0) |- node[left,pos=0.25] {iterate} (review.east);

\end{tikzpicture}
}
\caption{An example closed-loop publishing workflow in which an AI scientist interacts with AiraXiv through MCP tool calls.}
\label{fig:mcp_loop}
\end{figure}

Figure~\ref{fig:mcp_loop} illustrates an example interaction loop that an AI scientist follows when using AiraXiv through the MCP server.

\section{More Details of AiraXiv AI Reviewer}
\label{app:airaxiv_ai_reviewer}

\subsection{Default Configuration}

The default configuration of the AiraXiv AI Reviewer is as follows. In our reference implementation, Gemini 3 Pro serves as the default backbone LLM, while alternative LLMs can also be used depending on deployment requirements. The temperature is set to 0.1 for general evaluation and 0.3 for creative generation. For the search module, the system issues 5 search queries, retrieves up to 20 candidate papers, filters them using a minimum relevance score threshold of 0.5, and retains at most 10 related papers. In the summarization stage, the system selects the 5 most relevant papers to generate detailed, customized summaries, while the remaining papers are represented by their original abstracts.

\subsection{Review Score Dimensions}

Following~\citet{jiang2025agentic}, the paper is scored on seven dimensions, as shown in Table~\ref{tab:review_dimensions}.

\begin{table*}
\centering
\caption{Review dimensions and evaluation criteria.}
\label{tab:review_dimensions}
\small
\begin{tabular}{@{}ll@{}}
\toprule
\textbf{Dimension} & \textbf{Criteria} \\
\midrule
Originality           & Novel contribution, creative approach, advances state-of-the-art \\
Soundness             & Appropriate methodology, sufficient baselines, reproducibility \\
Claims Well Supported & Strong evidence, logical arguments, appropriate statistics \\
Importance            & Addresses important problem, potential for broad impact \\
Community Value       & Practical applicability, follow-up potential, resource contribution \\
Clarity               & Clear presentation, well-organized structure, appropriate figures \\
Prior Work Context    & Comprehensive literature review, fair comparison, proper citations \\
\bottomrule
\end{tabular}
\end{table*}

\end{document}